\documentclass[journal]{IEEEtran}
\usepackage{tabularx}
\usepackage{graphicx}
\usepackage{subfigure}  
\usepackage{adjustbox}
\usepackage{mathrsfs}
\usepackage{amsmath}
\usepackage{multirow}
\usepackage{mathrsfs}
\usepackage{algpseudocode}
\usepackage{amsmath}
\usepackage{amsfonts}
\usepackage{amssymb}
\usepackage{bbm}
\usepackage{algorithmicx,algorithm}

\begin{document}

\title{Rectified Meta-Learning from Noisy Labels for Robust Image-based Plant Disease Diagnosis}

\author{
        Ruifeng Shi,
        Deming Zhai,~\IEEEmembership{Member,~IEEE,}
        Xianming Liu,~\IEEEmembership{Member,~IEEE,}
       Junjun  Jiang,~\IEEEmembership{Member,~IEEE,} and~Wen Gao,~\IEEEmembership{Fellow,~IEEE}
\thanks{ R. Shi, D. Zhai, X. Liu, and J. Jiang are with School of Computer Science and Technology, Harbin Institute of Technology, Harbin, 150001, China; and also with Peng Cheng Laboratory, Shenzhen, China. e-mail: \{18S003075, zhaideming, csxm,  jiangjunjun\}@hit.edu.cn.

W. Gao is with School of Electrical Engineering and Computer Science,
Peking University, Beijing, 100871, China; and also with Peng Cheng Laboratory, Shenzhen, China. e-mail: wgao@pku.edu.cn.}}

\markboth{Journal of \LaTeX\ Class Files,~Vol.~14, No.~8, August~2015}%
{Shell \MakeLowercase{\textit{et al.}}: Rectified Meta-Learning from Noisy Labels for Robust Image-based Plant Disease Diagnosis}

\maketitle

\begin{abstract}
Plant diseases serve as one of main threats to food security and crop production. It is thus valuable to exploit recent advances of artificial intelligence to assist plant disease diagnosis. One popular approach is to transform this problem as a leaf image classification task, which can be then addressed by the powerful convolutional neural networks (CNNs). However, the performance of CNN-based classification approach depends on a large amount of high-quality manually labeled training data, which are inevitably introduced noise on labels in practice, leading to model overfitting and performance degradation. To overcome this problem, we propose a novel framework that incorporates rectified meta-learning module into common CNN paradigm to train a noise-robust deep network without using extra supervision information. The proposed method enjoys the following merits: i) A rectified meta-learning is designed to pay more attention to unbiased samples, leading to accelerated convergence and improved classification accuracy.  ii) Our method is free on assumption of label noise distribution, which works well on various kinds of noise. iii) Our method serves as a plug-and-play module, which can be embedded into any deep models optimized by gradient descent based method. Extensive experiments are conducted to demonstrate the superior performance of our algorithm over the state-of-the-arts.
\end{abstract}
\begin{IEEEkeywords}
Plant disease classification, robust deep learning, rectified meta-learning.
\end{IEEEkeywords}

\IEEEpeerreviewmaketitle

\section{Introduction}

\IEEEPARstart{D}{espite} remarkable progress has been made in the mass production and accessibility of food, plant diseases~\cite{StrangePlant2005} remain one of main threats to food security and crop production. In the developing countries, where the necessary infrastructures are lacking, more than 50\% of yield loss is due to pests and diseases~\cite{plant1}. The straightforward and effective way is to borrow the wisdom of agriculture experts to determine the type and severity of plant diseases. However, experts are very rare, and are also prone to misjudgments. 
\begin{figure}[t]
 \centering
 \includegraphics[width=0.45\textwidth]{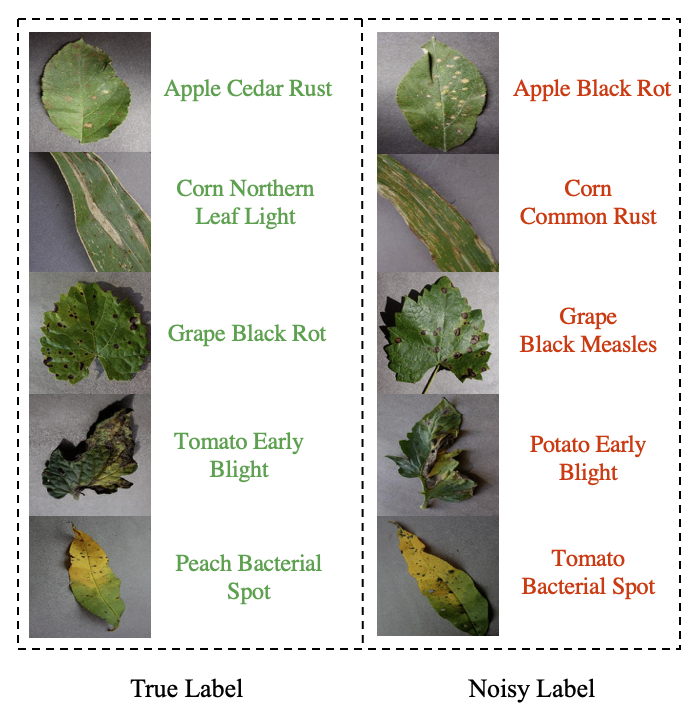}
 \caption{Examples of true and noisy labels in PlantVillage dataset.  The left side show images with true labels. The right side show images with noisy labels, whose labels should be ones of the left counterpart.}
 \label{fig:clean_noisy}
\end{figure}

In recent years, methods for automatically determining plant diseases by machine learning have received increasing attention~\cite{Khirade2015Plant,Waghmare2016Detection}. The basic idea of these learning-based methods is to study pictures of plant leaves and then predict types of plant diseases through image classification technique~\cite{FerentinosDeep,Kaur2016An}. Conventional methods complete this task based on handcrafted features. However, feature design is a non-trivial problem. With the advent of deep learning, the plant disease classification methods based on convolutional neural networks (CNNs)~\cite{Mohanty2016Using} have obtained better performance than traditional methods because of the powerful implicit feature extraction ability~\cite{GuanAutomatic2017,BarbedoFactors2018}. However, the performance of the CNN-based approach depends on a large amount of high-quality manually labeled training data, some of which are inevitably mislabeled in practice, as illustrated in Fig.~\ref{fig:clean_noisy}. The CNN-based approach lacks of the ability to distinguish clean labels and noisy labels, resulting in model overfitting and degraded performance. Therefore, how to design a robust method to handle label noise is critical to improve the performance of CNN-based plant disease diagnosis task.

To this end, various approaches attempted to design more complicated networks to estimate the noise~\cite{2016Jacob}, or correct the noisy labels by introducing a confusion matrix between true and noisy labels~\cite{DanUsing}. Most of these methods assume that a single mapping exists between true and noisy labels, and this mapping is independent of individual samples. However, in practical cases, the sources of noise on labels are diverse and the types of noise are complicated. It is difficult to employ a single mapping to simulate all scenarios. Furthermore, many methods tried to improve the performance by introducing additional supervision information~\cite{lee2017cleannet}. For example, some methods selected a part of training dataset and artificially correct the noisy labels by professional experts. By using the corrected training data, the derived deep networks surely can increase the ability of robustness to noise. However, these approaches are time-consuming and expensive when applied in large-scale real-world dataset. 
\begin{figure*}[t]
 \centering
 \includegraphics[width=1\textwidth]{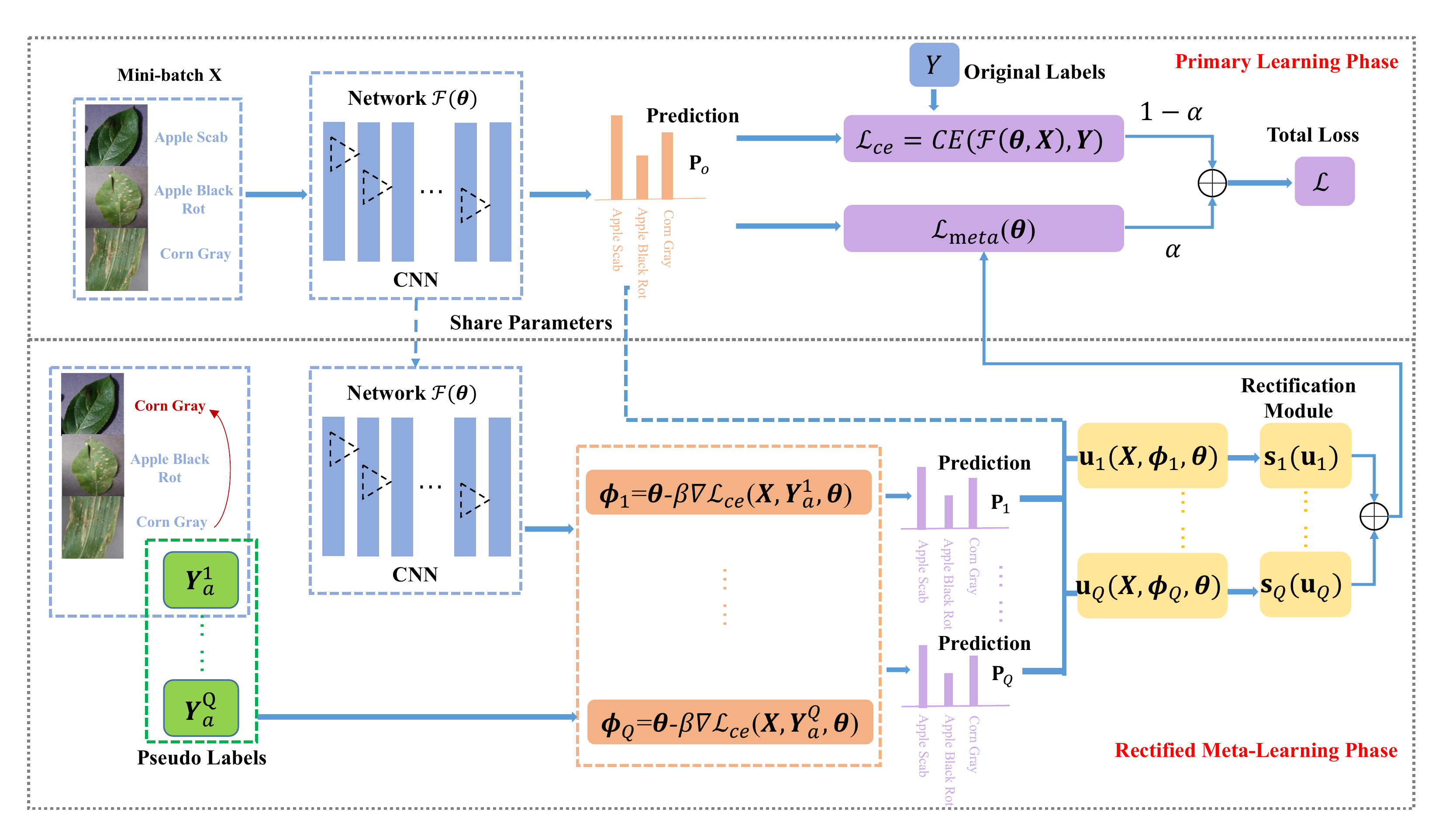}
 \caption{Illustration of the proposed noise-robust plant disease classification framework, which contains two phases that are performed iteratively: 1) primary learning phase, which is designed to train a network to predict the classification results according to original noisy labels; 2) rectified meta-learning phase, which is tailored to improve network robustness to label noise by self-learning. }
 \label{fig:framework}
\end{figure*}

In this study, we propose a novel framework that incorporates rectified meta-learning module into common CNN paradigm, which aims to train a noise-robust network for plant disease diagnosis without extra supervised information. As shown in Fig.~\ref{fig:framework}, our method contains two phases: the first phase (the upper part) is to train a CNN-based model to predict the classification results according to original noisy labels, and the second phase (the bottom part) is to combine meta-learning with rectification module to improve network tolerance of noise. Specifically, in the process of meta-learning, multiple synthetic min-batches with pseudo labels are generated to simulate the process of training with label noise. And then,
we train the network using one gradient update, and encourage it to give consistent predictions with that of the first phase. Furthermore, we propose a rectification module to give more penalty on highly-biased samples, while less penalty on unbiased ones, leading to less proneness to overfitting. 

The main contributions of this work are highlighted as follows:
\begin{itemize}
    \item [1)]
We propose a novel framework to reduce the impact of noisy labels on classification performance, without introducing additional manually-corrected information. Our method serves as a plug-and-play module, which can be applied on any deep models optimized by gradient descent based method, and does not rely on the assumption of label noise.  
    \item [2)] 
A rectification module is tailored into the network training to make it pay more attention to the unbiased samples, leading to accelerated convergence and improved classification accuracy.
    \item [3)] 
We conduct extensive experiments on public PlantVillage datasets~\cite{Hughes2015An} with various label noises, to demonstrate the superior performance of the proposed method over the state-of-the-arts.
\end{itemize}

The outline of the paper is as follows.
We overview related works in Section II.
In Section III, we elaborate the proposed  rectified meta-learning strategy for robust image-based plant disease classification.
We introduce the unified optimization framework in Section IV.
Extensive experimental results are provided in Section IV. Finally, conclusions are presented in Section V.

\section{Related Works}

In this section, we provide a brief overview about the related works, including learning with noisy labels and meta-learning.
\subsection{Learning with Noisy Labels}
Designing learning algorithms that can learn from data sets with noisy labels is
of great practical importance. Many works are presented to address this problem. One line of approaches use transition matrix~\cite{DanUsing,patrini2017making,Li2017Learning,Niu2018} to characterize the transition probability between noisy and true labels, \textit{i.e.}, each sample with a true label has a fixed probability of being mislabeled as a noisy label. For example, Patrini \textit{et al.}~\cite{patrini2017making} proposed a loss correction method to estimate the transition matrix by using a deep network trained on the noisy datasets. Hendrycks \textit{et al.}~\cite{DanUsing} exploited a subset of accurately labeled data to estimate the transition matrix. The above methods rely on a prior assumption that the transition matrix keeps identical between classes but is irrelevant to samples. Therefore, these methods worked well on the noisy datasets which is simple and artificial, such as the noisy CIFAR10~\cite{cifar10}. When faced with complex and real-world plant diseases datasets, such as PlantVillage dataset, these methods do not take effect since the above assumption is no longer reasonable. 

Another line of methods use a reference set to help training a robust network, which is a clean subset of
data including both noisy labels and verified true labels~\cite{lee2017cleannet,Scott2015,VeitLearning,Sukhbaatar2014Training}. For example, Lee \textit{i.e.} \cite{lee2017cleannet} introduced CleanNet, a joint neural embedding network, which requires a fraction of the classes being manually verified to provide the knowledge of label noise that can be transferred to other classes. Veit \textit{et al.} \cite{VeitLearning} proposed a multi-task network that jointly learns to clean noisy annotations and to accurately classify images. However, the construction of reference set is usually expensive and time-consuming. 

In addition, one popular line is to optimize the network and sample labels jointly~\cite{Wang2019,Tanaka2018Joint,han_deep_self_learning,Sheng2018CurriculumNet,DBLP}. For instance, in \cite{Tanaka2018Joint},  a joint optimization framework of learning DNN parameters and estimating true labels is proposed, which can correct labels during training by alternating update of network parameters and labels. Yi \textit{et al.} \cite{DBLP} proposed an end-to-end framework called PENCIL, which updates both network parameters and label estimations as label distributions. \cite{han_deep_self_learning} proposed a deep self-learning framework. By correcting the label using several class prototypes and training the network jointly using the corrected and original noisy iteratively, this work provides an effective end-to-end training framework without using an accessorial network or adding extra supervision on a real noisy dataset.

\subsection{Meta-Learning}
Meta-learning, or referred to as learning to learn, is an approach to train a model on a variety of learning tasks, such that it can learn and adapt quickly for new tasks using only a few samples \cite{Vilalta2001A,ChristianeMetalearning,Finn2017Model}. Meta-learning attracted considerable interest in the machine learning community in the last years, and many strategies have been proposed. 
The existing approaches in general can be divided into either memory-based, metric-based, or optimization-based methods. A memory-based method is proposed in~\cite{Adam2016}, which learns to store correct sample and label into the same memory slot and retrieve it later, in a task-generic manner. The metric-based approach attempts to learn a shared metric space~\cite{Snell2017}, which defines a distance between the sample and the class prototype, such that the samples are closer to their correct prototypes than to others.

In contrast to other two approaches, the optimization-based approach~\cite{ChenLearning,Finn2017Model,Dougal2015} is more popular, which turns to learn a shared initialization parameter that is optimal for any tasks within few gradient steps from the initial parameter. For example, Hochreiter \textit{et al.}~\cite{Hochreiter2001} introduced gradient descent method for meta-learning, specifically for recurrent meta-learning algorithms. This work is extended to reinforcement learning by Wang \textit{et al.}~\cite{Wang2016} and used for Recurrent Neural Network (RNN) by Andrychowicz \textit{et al.}~\cite{Andrychowicz2016}. The representative optimization-based method MAML~\cite{Finn2017Model} is model-agnostic and performs training by doing gradient updates on simulated meta-tasks, which aims to train model parameters that can learn well based on a few examples and a few gradient descent steps. Meta-SGD~\cite{Li2017} uses learning the learning rate differently for each parameter to enhance upon MAML. For effective learning of a meta-learner, meta-learning approaches adopt the episodic training strategy~\cite{Vinyals2016} that trains and evaluates a model over a large number of subtasks, which are called meta-training and meta-test phase, respectively. Our approach is similar to MAML. Both our approach and MAML are model-agnostic and perform training by doing gradient updates. However, MAML aims at few-shot transfer to new tasks, whereas our goal is to learn a noise-robust network for plant disease classification.

\section{Proposed Method}
In this section, we introduce in details the proposed rectified meta-learning scheme for plant disease classification. Firstly, we state the dilemma of learning from noisy labels. Then we detail the proposed methodology, which includes  two  phases---primary  learning  phase  and  rectified meta-learning phase---that are performed iteratively.
\subsection{Problem Statement}
In supervised plant disease classification problem, we have a set of $n$ training images $\boldsymbol{X}=[\boldsymbol{x}_{1},...,\boldsymbol{x}_{n}]$ with ground-truth labels $\boldsymbol{Y}=[\boldsymbol{y}_{1},...,\boldsymbol{y}_{n}]$, where $\boldsymbol{x}_{i}$ denotes the $i^{th}$ sample and $\boldsymbol{y}_{i}\in \left \{ 0,1 \right \}^{c}$ is a one-hot vector representing the corresponding noisy label over $\emph{c}$ classes. A neural network $\mathcal{F}(\boldsymbol{\theta})$ with parameter $\boldsymbol{\theta}$ is defined to map image $\boldsymbol{x_{i}}$ to label probability distribution $\mathcal{F} \left ( \boldsymbol{x}_i,\boldsymbol{\theta} \right)$. When training on a truly-labeled dataset, the optimization problem is defined as follows:
\begin{equation}
\boldsymbol{\theta ^*}= \arg\min_{\boldsymbol{\theta}} \mathcal{L} \left (\mathcal{F}\left ( \boldsymbol{X},\boldsymbol{\theta} \right ), \boldsymbol{Y} \right ),
\end{equation}
where $\mathcal{L}$ represents the empirical risk. However, in practice, the labels $\boldsymbol{Y}$ inevitably contain some noise, \textit{i.e.}, some image samples are annotated with inaccurate labels, making this end-to-end approach sub-optimal. The network would overfit to label noise and suffer from classification performance degradation.

In view of the above limitation, we propose a two-phase approach to learn a noise-robust deep network for plant disease classification. The first phase is the common CNN-based classification process, where we introduce the meta-learning loss derived from the second phase into its loss function to promote the network less prone to label noise.
In the second phase,  the meta-learning process generates multiple synthetic min-batches with pseudo labels, according to which we simulate the process of training with label noise. We then seek the network parameters that are less sensitive to label noise in the parameter space by minimizing the weighted consistency loss among network outputs. In this way, we can learn the underlying reliable knowledge from data with label noise. In contrast, the most recent method \cite{han_deep_self_learning} adopts a cascade manner, which first corrects the noisy labels and then learns the network parameters.




\subsection{Methodology}
The overall framework of our method is illustrated in Fig.~\ref{fig:framework}. It contains two phases: primary learning phase and rectified meta-learning phase, which are performed iteratively.


\subsubsection{Primary Learning Phase}
The pipeline of the primary learning phase is illustrated in the upper part of Fig.~\ref{fig:framework}. It aims to optimize the parameters $\boldsymbol{\theta}$ of the deep neural network $\mathcal{F}$, according to which the classification of plant disease is conducted.  At the beginning, we collect a mini-batch data $\left ( \boldsymbol{X},\boldsymbol{Y} \right )$\footnote{To avoid introducing more notations, here we re-use ($\boldsymbol{X},\boldsymbol{Y} )$ to denote a mini-batch.} from the training set, where $\boldsymbol{X}=\left ( \boldsymbol{x}_{1},...,\boldsymbol{x}_{k} \right )$ are $k$ image samples, and $\boldsymbol{Y}=\left ( \boldsymbol{y}_{1},...,\boldsymbol{y}_{k} \right )$ are the corresponding labels which contain noise.
$\boldsymbol{X}$ is then taken as the input of $\mathcal{F}$ to produce the corresponding label prediction $\mathcal{F} \left ( \boldsymbol{ X}, \boldsymbol{\theta} \right)$. The loss function of the network training consists of two parts: 
\begin{itemize}
\item The first part is the cross-entropy loss $\mathcal{L}_{ce}$ that is defined as follows:
\begin{equation}
\begin{aligned}
\mathcal{L}_{ce}(\boldsymbol{\theta}) &=CE(\mathcal{F}\left ( \boldsymbol{X},\boldsymbol{\theta} \right ),\boldsymbol{Y} )
\\&=-\frac{1}{n}\sum_{i=1}^{n}\sum_{j=1}^{c}\boldsymbol{y}_{ij}\cdot \log f_{j}\left ( \boldsymbol{ x}_{i},\theta \right ),
\end{aligned}
\end{equation}
where $f$ maps an input to an output of the $c$-class softmax layer. 
\item The second part is the meta-learning loss $\mathcal{L}_{meta}$, which is added  to make network more robust to label noise. 

\item The total loss function is then defined as follows:
\begin{equation}
\mathcal{L}(\boldsymbol{\theta})=(1-\alpha )\mathcal{L}_{ce}(\boldsymbol{\theta})+\alpha\mathcal{L}_{meta}(\boldsymbol{\theta}),
\label{eq:alpha}
\end{equation}
where $\alpha \in \left [ 0,1 \right ]$ is the weight factor that compromises contributions of the two losses. At the very beginning of iteration, $\alpha$ is set to 0 and the network $\mathcal{F}$ is trained only using the cross-entropy loss $\mathcal{L}_{ce}$. In the rest iterations, the meta-learning loss joins into the network training, where $\alpha$ is set as a positive value. 
\end{itemize}

The total loss function can be addressed by stochastic gradient descent (SGD). The model parameters $\boldsymbol{\theta}$ are updated as follows:
\begin{equation}
\boldsymbol{\theta} \leftarrow \boldsymbol{\theta} -\gamma \nabla_{\boldsymbol{\theta}} \left ( \left ( 1-\alpha  \right )\mathcal{L}_{ce}(\boldsymbol{\theta}) + \alpha \mathcal{L}_{meta}(\boldsymbol{\theta}) \right ),
\end{equation}
where $\gamma$ is the learning rate. The full algorithm is outlined in Algorithm 1.

\subsubsection{Rectified Meta-Learning Phase}
The pipeline of the rectified meta-learning phase is illustrated in the bottom half of Fig. 2.  The rectified meta-learning phase is tailored to make the primary learning phase less prone to overfitting to label noise. Using multiple synthetic mini-batches, we train the network using one gradient update, and enforce it to give consistent predictions with that of primary learning phase. A rectification module is further designed to pay more attention to the unbiased samples, leading to accelerated convergence and improved classification accuracy.

\begin{itemize}
\item
    \textbf{Meta-Learning}
\end{itemize}

To perform meta-learning, for each input mini-batch, we first generate multiple synthetic mini-batch with pseudo labels $\left \{ \hat{\boldsymbol{Y}}^{1},...,\hat{\boldsymbol{Y}}^{Q} \right \}$ that are from a similar distribution as the original noisy labels. The generation process is described as follows \cite{Li2018Learning}.
Firstly, we randomly select $m$ samples from $k$ ones in each mini-batch $\boldsymbol{X}$, which are denoted as $\boldsymbol{X}_{m}=(\boldsymbol{x}_{1},...,\boldsymbol{x}_{m})$. For each element $\boldsymbol{x}_i$ in $\boldsymbol{X}_{m}$, we find its top 8 nearest neighbors in  $\boldsymbol{X}$ and randomly choose one of them, \textit{e.g.}, $\boldsymbol{x}_j$. Then we replace the label of $\boldsymbol{x}_i$ as that of $\boldsymbol{x}_j$, \textit{i.e.}, $\boldsymbol{y}_{j}$.
The above procedure is repeated for $Q$ times to get $Q$ sets of pseudo labels. $Q$ is empirically set as $10$.

With the generated pseudo labels, we define a set of synthetic mini-batch as $\{\left ( \boldsymbol{X},\hat{\boldsymbol{Y}}^{i} \right )\}_{i=1}^Q$. For each synthetic mini-batch, we update the network parameter $\boldsymbol{\theta}$ to $\boldsymbol{\phi} _{i}$ using one gradient descent step:
\begin{equation}
\boldsymbol{\phi} _{i}=\boldsymbol{\theta} - \beta \nabla_{\boldsymbol{\theta}} \mathcal{L}_{ce}\left (\mathcal{F}( \boldsymbol{X},\boldsymbol{\theta}),\hat{\boldsymbol{Y}}^{i} \right ),
\end{equation}
where $\boldsymbol{\theta}$ denotes the network parameter obtained in primary learning phase, $\beta$ denotes the meta learning rate. According to the updated network, we get the predicted distribution $\{\boldsymbol{P}_{i}\}_{i=1}^Q$.

To promote the deep model to be more tolerant to noisy label, we enforce outputs of the updated network to be consistent with the prediction $\boldsymbol{P}_{o}$ of the primary learning phase. Specifically, 
we measure the consistency between $\boldsymbol{P}_{o}$ and $\boldsymbol{P}_{i}$ by the Kullback-Leibler (KL) divergence:
\begin{equation}
\small
\begin{aligned}
\boldsymbol{u}_{i}\left ( \boldsymbol{X}, \boldsymbol{\phi} _{i} , \boldsymbol{\theta} \right )&=  \mathcal{D}_{KL} \left ( \boldsymbol{P}_{o} || \boldsymbol{P}_{i} \right )\\&=\mathcal{D}_{KL}\left ( f\left ( \boldsymbol{X},\boldsymbol{\theta}  \right ) || f\left ( \boldsymbol{X},\boldsymbol{\phi}_{i}  \right ) \right )\\
&=\sum_{i=1}^{n} f\left ( \boldsymbol{x}_{i},\boldsymbol{\theta} \right ) \cdot \log\frac{f\left ( \boldsymbol{x}_{i},\boldsymbol{\theta} \right )}{f\left ( \boldsymbol{x}_{i},\boldsymbol{\phi} _{i}  \right )}.
\end{aligned}
\end{equation}

\begin{itemize}
\item\textbf{Rectification Module}
\end{itemize}
\begin{figure}[t]
\centering
\subfigure{
\begin{minipage}[t]{1.0\linewidth}
\centering
\tiny
\includegraphics[width=0.7\textwidth]{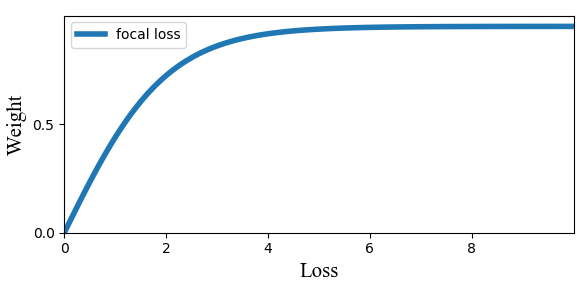}
\centerline{(a) The weighting function of the focal loss}
\end{minipage}%
}%
\\
\subfigure{
\begin{minipage}[t]{1.0\linewidth}
\centering
\tiny
\includegraphics[width=0.7\textwidth]{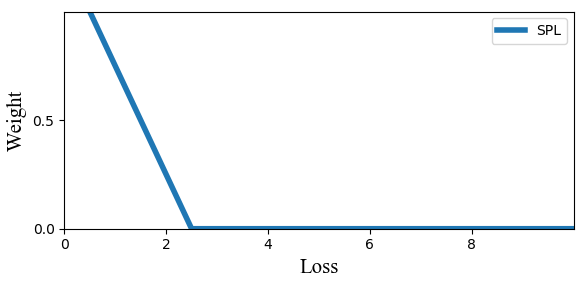}
\centerline{(b) The weighting function of self-paced learning (SPL)}
\end{minipage}%
}%
\begin{center}
\caption{Illustration of two representative weighting functions.}
\label{figure:result}
\end{center}
\end{figure}

With the above consistency loss, a straightforward idea is to define the meta-learning loss $\mathcal{L}_{meta}$ as the summed consistency loss of all $Q$ updated models with parameters $\{\boldsymbol{\phi}_i\}_{i=1}^Q$, as done in \cite{Li2018Learning}:
\begin{equation}
\begin{aligned}
\mathcal{L}_{meta}(\boldsymbol{\theta})=\frac{1}{Q}\sum_{i=1}^Q \boldsymbol{u}_{i}\left ( \boldsymbol{X}, \boldsymbol{\phi} _{i} , \boldsymbol{\theta} \right ).
\end{aligned}
\end{equation}

However, this approach implies that each synthetic mini-batch contributes equally to the final decision. It is unreasonable since the pseudo labels produce more or less noise, leading to various confidence of $Q$ synthetic mini-batches.
In our work, we propose a rectification module, which works like the attention mechanism by assigning different weights to mini-batches. 

The rest critical problem is how to design an effective weighting function. In the literature, there exist two entirely contradictive ideas for constructing such a weighting function. One kind of methods make the weight function monotonically increasing, as depicted in Fig.~\ref{figure:result} (a), \textit{i.e.}, enforce the learning process to more emphasize samples with larger loss since they are more likely with right labels. Representative methods include focal loss~\cite{Lin2017Focal} and hard negative mining~\cite{Malisiewicz2011Ensemble}. In contrast, the other kind of methods set the weighting function monotonically decreasing, as shown in Fig.~\ref{figure:result}  (b), to suppress effects of samples with extremely large loss values, which are possible with wrong labels. Typical methods of this category include self-paced learning (SPL~\cite{Kumar2010Self}) and iterative re-weighting~\cite{ZhangGeneralized}.

\begin{figure}[t]
\begin{minipage}[b]{1.0\linewidth}
  \centering
  \centerline{\includegraphics[width=1.0\textwidth]{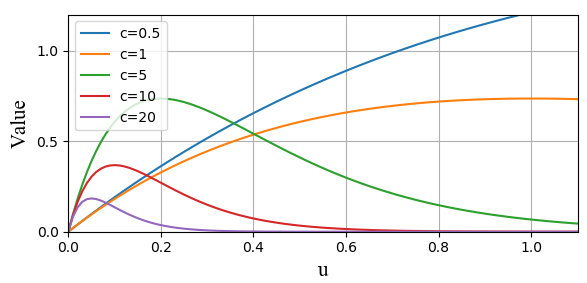}}
  \caption{Illustration of our proposed weighting function. The parameter $c$ controls the shape of function.}\label{fig:rectification}
\end{minipage}
\end{figure}

However, for our task, these two weighting functions are not feasible. On one hand, if we follow the first approach that more cares mini-batches with large consistency loss, which means the produced pseudo labels are close to the noisy ones, the rectified meta-learning phase would not be helpful to make the primary learning phase robust against label noise. On the other hand, if we follow the second approach that emphasizes more mini-batches with small consistency loss, which means that the produced pseudo labels are wrong, the ovefitting problem of the network would be even worse. Therefore, by combining these two methods, we propose a rectification module that increases the penalty of small and large consistency losses and reduces the penalty of moderate ones. Specifically, we define the rectification function $\boldsymbol{s}_{i}$ as:
\begin{equation}
\boldsymbol{s}_{i}\left ( \boldsymbol{u} _{i} \right )=\boldsymbol{u}_{i}\cdot \exp\left ( -c\cdot \boldsymbol{u}_{i} \right ),
\end{equation}
where $c$ is the parameter controlling the shape of function.
Fig.~\ref{fig:rectification} depicts the function behavior of $\boldsymbol{s}_{i}$. It can be seen that the penalty first goes up but then drops towards $0$ as $\boldsymbol{u}_{i}$ increases. 

Finally, we define meta-learning loss as a weighted average of all consistency losses:
\begin{equation}
\small
\begin{aligned}
\mathcal{L}_{meta}(\boldsymbol{\theta})&=\frac{1}{Q} \sum_{i=1}^{Q} \boldsymbol{s}_{i}\left ( \boldsymbol{u} _{i} (\boldsymbol{X}, \boldsymbol{\phi} _{i} , \boldsymbol{\theta}) \right )\\
&=\frac{1}{Q} \sum_{i=1}^{Q} \boldsymbol{s}_{i}\left ( \boldsymbol{u} _{i} (\boldsymbol{X},\boldsymbol{\theta} - \beta \nabla_{\boldsymbol{\theta}}  \mathcal{L}_{ce}\left ( \mathcal{F}(\boldsymbol{X,\theta}),\hat{\boldsymbol{Y}}^{i}  \right ),\boldsymbol{\theta}) \right ).
\end{aligned}
\end{equation}

\begin{algorithm}[tb]
\caption{Noise-robust Classification Algorithm }
\label{alg:algorithm}
\begin{algorithmic}[1] 
\State Randomly initialize $\boldsymbol{\theta}$.
\For{$M=1:$ num\underline{~}epochs}
\If{$M<$start\underline{~}epoch}
\State Sample $(\boldsymbol{X},\boldsymbol{Y})$ from training datasets.
\State update $\boldsymbol{\theta} \leftarrow \boldsymbol{\theta} -\gamma \nabla_{\boldsymbol{\theta}} \mathcal{L}_{ce}\left ( \mathcal{F}(\boldsymbol{X},\boldsymbol{\theta}),\boldsymbol{Y}  \right )$
\Else
\State Sample a mini\underline{~}batch $(\boldsymbol{X},\boldsymbol{Y})$ of size $k$ from training datasets.
\For{$i=1:Q$}
\State Generate pseudo labels $\hat{\boldsymbol{Y}}^{i}$.
\State Compute updated parameters with gradient descent$:\boldsymbol{\phi} _{i}=\boldsymbol{\theta} - \beta \nabla_{\boldsymbol{\theta}} \mathcal{L}_{ce}\left ( \mathcal{F}(\boldsymbol{X},\boldsymbol{\theta}),\hat{\boldsymbol{Y}}^{i}  \right )$
\State Evaluate consistency loss$:\boldsymbol{u}_{i}\left (\boldsymbol{X}, \boldsymbol{\phi} _{i} , \boldsymbol{\theta} \right )=\mathcal{D}_{KL}\left ( f\left ( \boldsymbol{X},\boldsymbol{\theta}  \right ) || f\left ( \boldsymbol{X},\boldsymbol{\phi} _{i}  \right ) \right )$
\State Compute the rectification module$:\boldsymbol{s}_{i}\left ( \boldsymbol{u}_{i} \right )=\boldsymbol{u}_{i}\cdot \exp\left ( -c\cdot \boldsymbol{u}_{i} \right )$
\EndFor
\State Evalute $\mathcal{L}_{meta}(\boldsymbol{\theta})=\frac{1}{Q} \sum_{i=1}^{Q} \boldsymbol{s}_{i}\left ( \boldsymbol{u} _{i}(\boldsymbol{X}, \boldsymbol{\phi} _{i} , \boldsymbol{\theta}) \right )$
\State $\boldsymbol{\theta} \leftarrow \boldsymbol{\theta} -\gamma \nabla_{\boldsymbol{\theta}} \left ( \left ( 1-\alpha  \right )\mathcal{L}_{ce}(\boldsymbol{\theta}) + \alpha \mathcal{L}_{meta}(\boldsymbol{\theta}) \right )$
\EndIf
\EndFor

\end{algorithmic}
\end{algorithm}

\section{Experiments}
In this section, extensive experimental results are provided to demonstrate the superior performance of our proposed method. We also offer comprehensive ablation study to help deep understanding about our scheme.
\subsection{Datasets}
\begin{table*}
\setlength\extrarowheight{1pt}
\caption{Classification accuracy ($\%$) of compared methods under symmetric label noise.}
\centering
\setlength{\tabcolsep}{5mm}{
\begin{tabular}{|l|l|l|l|l|l|l|l}
\hline
Method & $\rho $ = 0 & $\rho $ = 0.1 & $\rho $ = 0.3 & $\rho $ = 0.5 & $\rho $ = 0.7& $\rho $ = 0.9\\ 
\hline
Cross-Entropy& 98.42& 98.09 & 95.17 &93.30 &79.27 & 38.36\\ 
Mohanty's    & 98.45& 98.25 & 95.78 &93.44 &79.27 & 40.18\\ 
Joint Optimization  & 98.27& 98.49 & 97.14 &95.78 &83.49 & 61.81\\ 
MLNT& 98.56& 98.41 & 97.61 &96.17 &85.96 & 63.04\\ 
Our Model    & \textbf{99.52}& \textbf{99.04} & \textbf{98.25} &\textbf{96.97} &\textbf{87.56} & \textbf{64.41}\\ 
\hline
\end{tabular}}
\label{tab:symmetric}
\end{table*}

\begin{table*}
\setlength\extrarowheight{1pt}
\caption{Classification accuracy ($\%$) of compared methods under asymmetric label noise.}
\centering
\setlength{\tabcolsep}{5mm}{
\begin{tabular}{|l|l|l|l|l|l|l|l}
\hline
Method & $\rho $ = 0.1 & $\rho $ = 0.2 & $\rho $ = 0.3 & $\rho $ = 0.4 & $\rho $ = 0.5\\ 
\hline
Cross-Entropy& 98.25& 96.45 & 91.39 &84.53 &77.51 \\ 
Mohanty's    & 98.25& 96.72 &91.24  &84.74 &77.71 \\ 
Joint Optimization       & 98.47& 96.66 & 94.21 &91.53 &79.94 \\ 
MLNT         & 98.09& 96.81 & 95.53 &92.95 &80.38 \\ 
Our Model   & \textbf{99.04}& \textbf{97.61} & \textbf{96.33} &\textbf{94.10} &\textbf{82.14} \\ 
\hline
\end{tabular}}
\label{tab:asymmetric}
\end{table*}

\begin{table*}
\setlength\extrarowheight{1pt}
\caption{Classification accuracy ($\%$) of compared methods under mixed label noise.}
\centering
\setlength{\tabcolsep}{5mm}{
\begin{tabular}{|l|l|l|l|l|l|l|l}
\hline
Method & $\rho $ = 0.1 & $\rho $ = 0.2 & $\rho $ = 0.3 & $\rho $ = 0.4 & $\rho $ = 0.5\\ 
\hline
Cross-Entropy& 93.14 &91.86  &88.20 &82.93 & 74.82\\ 
Mohanty's    & 94.12 &92.97  &88.42 &83.09 & 73.68\\ 
Joint Optimization       & 95.61 & 95.14 &94.72 &89.70& 78.23\\
MLNT         & 98.41 & 98.15 &97.09 &94.12& 81.39 \\ 
Our Model    & \textbf{98.72} & \textbf{98.41} & \textbf{97.93} & \textbf{94.58}& \textbf{83.14}\\ 
\hline
\end{tabular}}
\label{tab:mixed}
\end{table*}

We conduct experimental comparison on 30 healthy and diseased plants from the well-known PlantVillage~\cite{Hughes2015An}, including apples, potatoes, corn, etc. Since the selected images are all annotated by professional experts, it can be seen as a clean dataset. We randomly select 10$\%$ of the dataset for validation, and manually corrupt the rest with three types of label noise: 
\begin{itemize}
\item \textbf{Symmetric Noise (SN)}, which is is injected by using a random one-hot vector to replace the ground-truth label of a sample with a probability of $\rho$~\cite{Li2018Learning}.
\item \textbf{Asymmetric Noise (AN)}, which is designed to mimic structure of real mistakes for similar classes~\cite{Li2018Learning}. For example, 
APPLE VENTURIA INAEQUALIS $\leftrightarrow $ APPLE CEDAR RUST, CORN GRAY LEAF SPOT $\leftrightarrow $ CORN COMMON RUST, 
GRAPE BLACK ROT $\leftrightarrow $ GRAPE BLACK MEASLES, POTATO EARLY BLIGHT $\leftrightarrow $ POTATO LATE BLIGHT, 
etc. These transitions are parameterized by $\rho  \in \left [ 0,1 \right ]$.
\item \textbf{Mixed Noise (MN)}, which is injected including both symmetric and asymmetric label noise to simulate more complicated real-world cases. For each sample, we randomly (with the probability of 50\%) choose a noise type to be added. The parameter $\rho $ controls noise level of chosen noise type.
\end{itemize}

\subsection{Experimental Setup}
The main structure of our network follows ResNet-50. We normalize the images, and perform data augmentation by random horizontal flip and random crop a $224\times 224$ patch from the resized image. We use a batch size $k=64$ with the learning rate of $\gamma=0.2$, and update $\boldsymbol{\theta}$ using SGD with a momentum of $0.9$ and a weight decay of $10^{-4}$. The total training processes contain 120 epochs. The learning rate is decreased by a factor of 10 at each 40 epochs. 

\subsection{Comparison with the state-of-the-arts}
We compare the proposed method with the state-of-the arts under symmetric, asymmetric and mixed label noise with various noise levels $\rho$. The comparison study includes:
\begin{itemize}
\item \textbf{Cross-Entropy:} It actually is the primary learning phase of our scheme without the meta-learning loss.  
\item \textbf{Mohanty~\cite{Mohanty2016Using}: } It is a state-of-the-art CNN-based plant disease classification method without considering the influence of noisy label.
\item \textbf{Joint Optimization~\cite{Tanaka2018Joint}: } It is a state-of-the-art method by alternatively updating the network parameters and correcting noisy labels during training. 
\item \textbf{MLNT~\cite{Li2018Learning}: } It performs meta-learning on noisy labels, which is most related to our method. We show its best performance in comparison.
\end{itemize}
The codes of these compared methods are publicly available. For the very recent method \cite{han_deep_self_learning}, there is no source code available, preventing us from comparing with it.

Table \ref{tab:symmetric} and \ref{tab:asymmetric}  show the comparison results under symmetric and asymmetric label noise, respectively. It can be found that, our method consistently outperforms the others, no matter what kind of noise and noise level are. Specifically, our scheme significantly improves the classification accuracy over Cross-Entropy, which demonstrates the effectiveness of the proposed meta-learning loss. Our scheme also achieves higher accuracy than Mohanty's method~\cite{Mohanty2016Using}, which demonstrates the effectiveness of our self-learning strategy from noisy label.  Cross-Entropy and the Mohanty's method \cite{Mohanty2016Using} do not consider the noisy label. In Table \ref{tab:symmetric}, in the case that the noise ratio $\rho $ is 0, which means there is no noisy labels, our method still works better than them. This shows our method has a good generalization ability.

Our method is superior to Joint Optimization and MLNT, which are state-of-the-art works that learn from noisy labels. Compared to Joint Optimization, our method achieves up to 4.07\% improvement of in accuracy. In addition, Joint Optimization method requires the ground-truth class distribution among training data, while our method does not require any prior knowledge or extra information about the dataset. When compared with MLNT method, our method achieves up to 1.76\% improvement in accuracy, which demonstrates the effectiveness of the proposed rectification module.


To demonstrate the effectiveness of the proposed method on complicated label noise, we further report the comparison results under mixed label noise in Table \ref{tab:mixed}. Our method outperforms the baseline methods---Cross-Entropy and  \cite{Mohanty2016Using}---by a large margin, \textit{e.g.}, improving the average accuracy from $89.03\%$ and $89.65\%$ to $97.41\%$. The larger the noise level, the more performance improvement is brought by our rectified meta-learning strategy. When compared with the two state-of-the-art methods, Joint Optimization and MLNT, the gains of our method are up to 4.88\% and 0.85\%, respectively.
These results verify that the proposed rectified meta-learning module indeed helps to learn the underlying knowledge from data in the present of label noise.

\begin{table}[t]
\setlength\tabcolsep{4pt}
\setlength\extrarowheight{3pt}
\caption{Overall label accuracy of the labels in original mixed noisy dataset with noise level $\rho = 0.4$ (Original), accuracy of the pseudo labels generated by the rectified meta-learning phase in first iterative cycle (Correct Initial) and accuracy of the pseudo labels generated by the final model when training ends (Correct Final).}
\centering
\setlength{\tabcolsep}{3mm}{
\begin{tabular}{c|c|c|c}
\hline
 & Original & Correct Initial & Correct Final \\ 
\hline
Accuracy &60.00\%  &88.79\% & \textbf{94.58\%}\\ 
 \hline
\end{tabular}}
\label{tab:classification_acc}
\end{table}

\begin{figure}[t]
 \centering
 \includegraphics[width=0.45\textwidth]{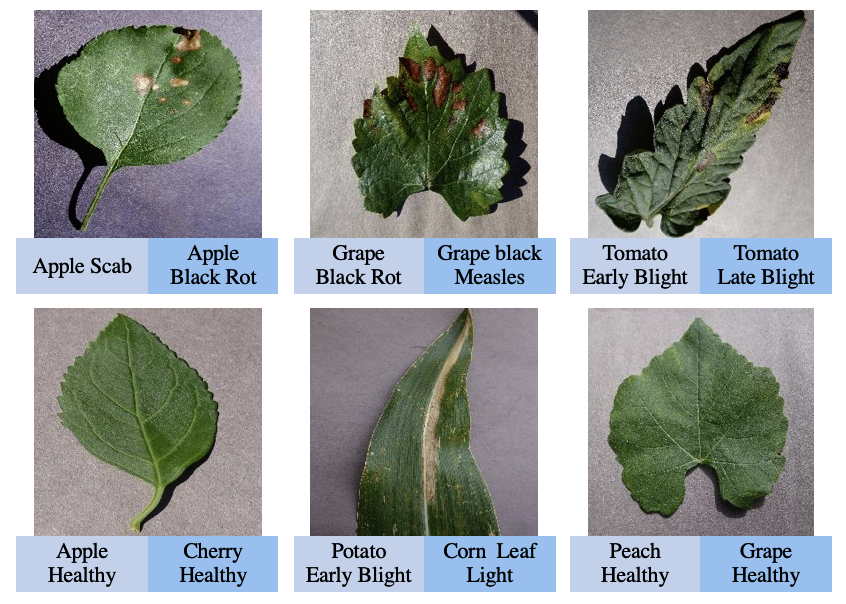}
 \caption{\textbf{Left}: The original noisy labels. \textbf{Right}: The corrected pseudo labels by our method.}
 \label{fig:pseudo_labels}
\end{figure}

\subsection{Ablation Study}
As  presented in the earlier part of this paper, our main contribution is the rectified meta-learning phase, which serves as a plug-and-play module that can be embedded into the primary learning phase. To demonstrate its effectiveness, we provide ablation study on label correction accuracy and convergence speed. Moreover, we investigate the influence of two important hyper-parameters---$Q$ that is the number of synthetic mini-batches and the weight factor $\alpha$---to the final classification performance. 

\subsubsection{The Influence of Rectified Meta-learning to Label Correction Accuracy}

We take the case under mixed noise (MN) with noise level $\rho = 0.4$ as an example to investigate the influence of the proposed rectified meta-learning to label correction accuracy.
As shown in Table \ref{tab:classification_acc}, the label accuracy of the original noisy dataset is 60\%, which is improved to 88.79$\%$ after the initial iterative cycle of the preliminary learning phase. When the proposed rectified meta-learning phase is incorporated, \textit{i.e}, learning with the entire network shown in Fig.~\ref{fig:framework}, the accuracy is further increased to 94.58$\%$, with 34.58\% and 5.79\% improvement compared with original dataset and the preliminary learning phase, respectively. 

We further explore the classification accuracy for all 30 classes as shown in Fig. \ref{fig:label_acc}. We can find that the original accuracies of all classes is around 60$\%$. For most classes (25 out of 30), our approach can increase the label accuracy to higher than 92$\%$. Specifically, for the $11^{th}$ class (``Grape Black Measles”) that is with about 66.13$\%$ original label accuracy, our scheme can improves the accuracy by 31.80$\%$ where 15.78\% is attributed to the proposed rectified meta-learning phase. Some of the noisy samples are successfully corrected by our approach, as shown in Fig. \ref{fig:pseudo_labels}. The above study demonstrates that the proposed rectified meta-learning phase is indeed helpful for noisy labels correction.

\begin{figure}[t]
\centering
\subfigure{
\begin{minipage}[t]{1.0\linewidth}
\centering
\tiny
\includegraphics[width=0.7\textwidth]{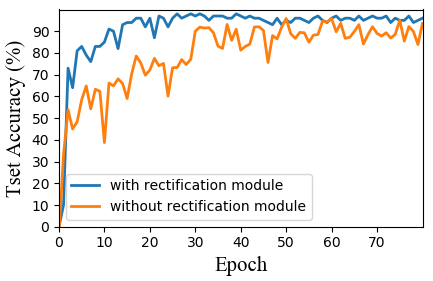}
\centerline{(a) mixed noise with noise level $\rho = 0.2$}
\end{minipage}%
}%
\\
\subfigure{
\begin{minipage}[t]{1.0\linewidth}
\centering
\tiny
\includegraphics[width=0.7\textwidth]{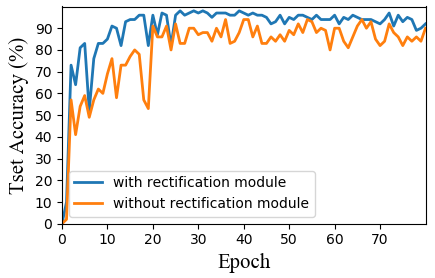}
\centerline{(b) mixed noise with noise level $\rho = 0.3$}
\end{minipage}%
}%
\caption{Illustration of the influence of the proposed rectification module. The rectification module is helpful for accelerating convergence and improving the classification accuracy.}
\label{fig:Progressive}
\end{figure}

\begin{figure*}[t]
 \centering
 \includegraphics[width=1\textwidth]{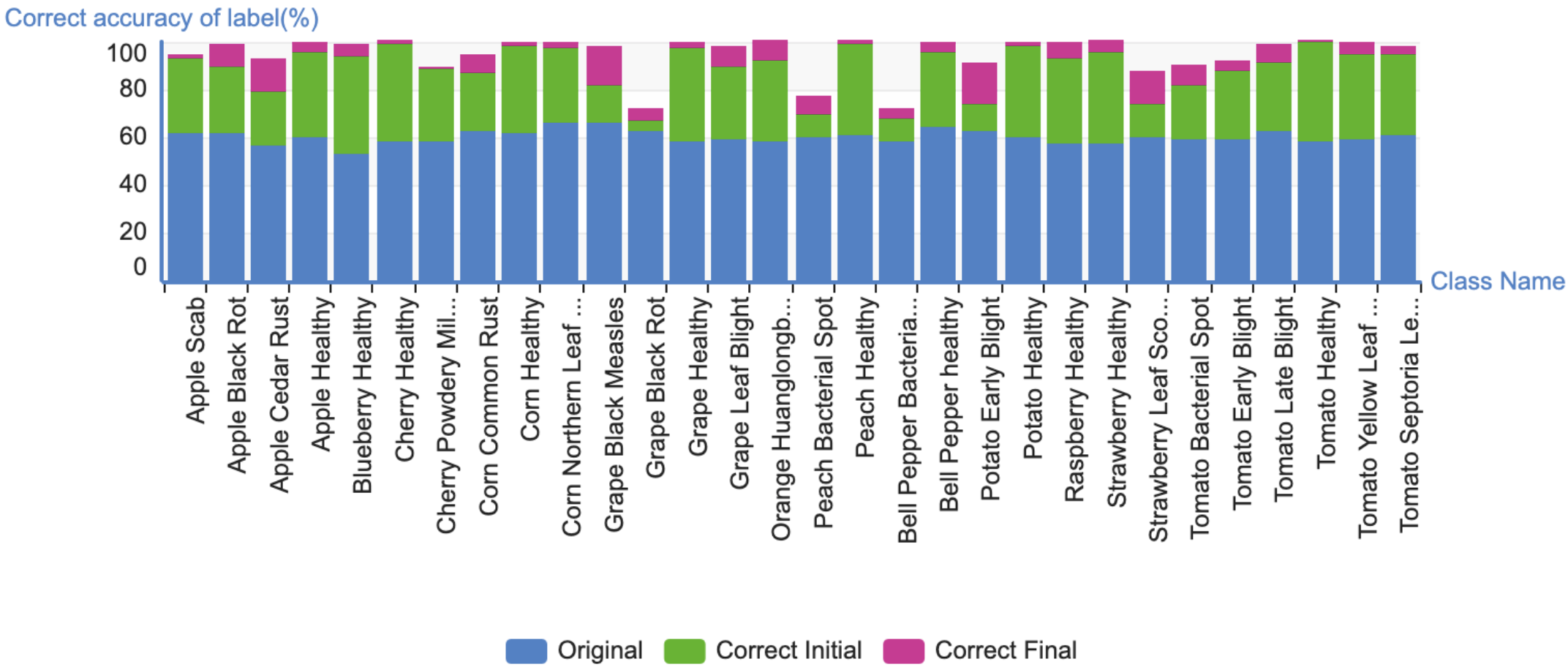}
 \caption{The label accuracy ($\%$) of labels in the original noisy dataset (Original), pseudo labels by the rectified meta-learning phase in the first iterative cycle (Correct Initial) and pseudo labels by the final model(Correct Final) for each class in PlantVillag}
 \label{fig:label_acc}
\end{figure*}

\begin{figure}[t]
\begin{minipage}[b]{1.0\linewidth}
  \centering
  \centerline{\includegraphics[width=1.0\textwidth]{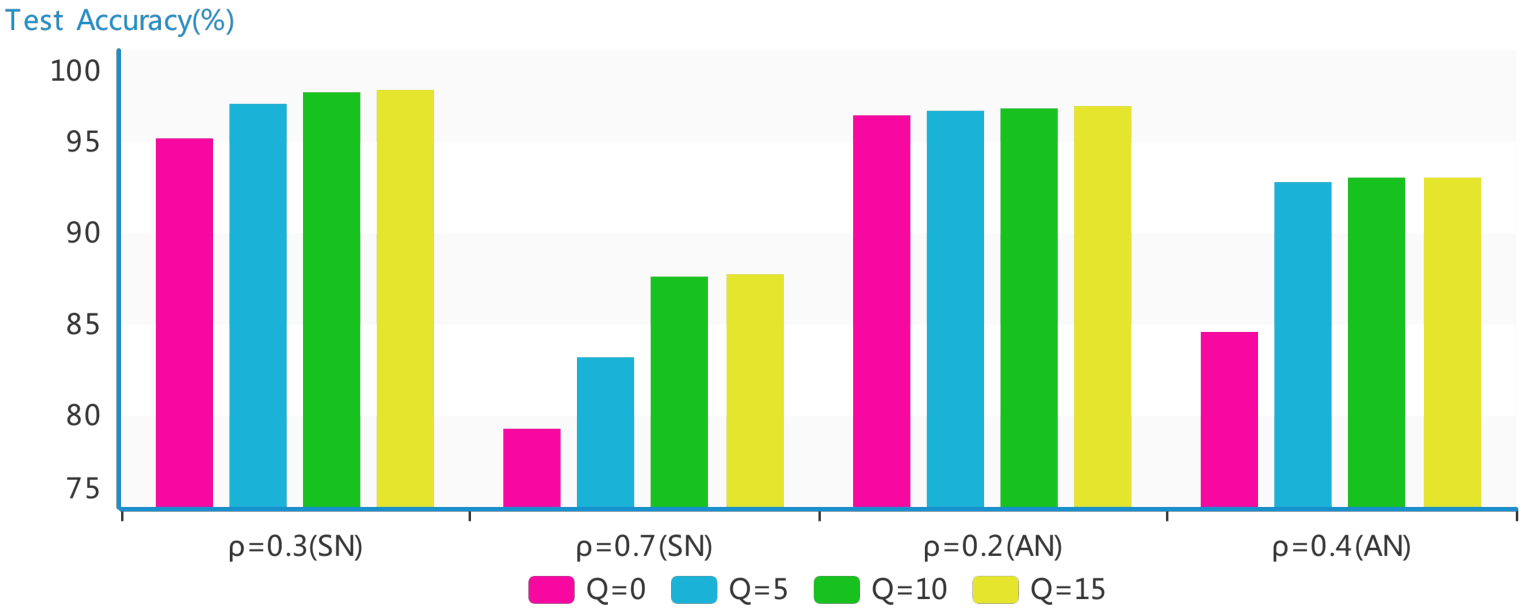}}
  \caption{Illustration of the influence of the number of synthetic mini-batches $Q$ to classification accuracy. }
  \label{fig:parameter q}
\end{minipage}
\end{figure}

\begin{figure}[t]
\begin{minipage}[b]{1.0\linewidth}
  \centering
  \centerline{\includegraphics[width=1.0\textwidth]{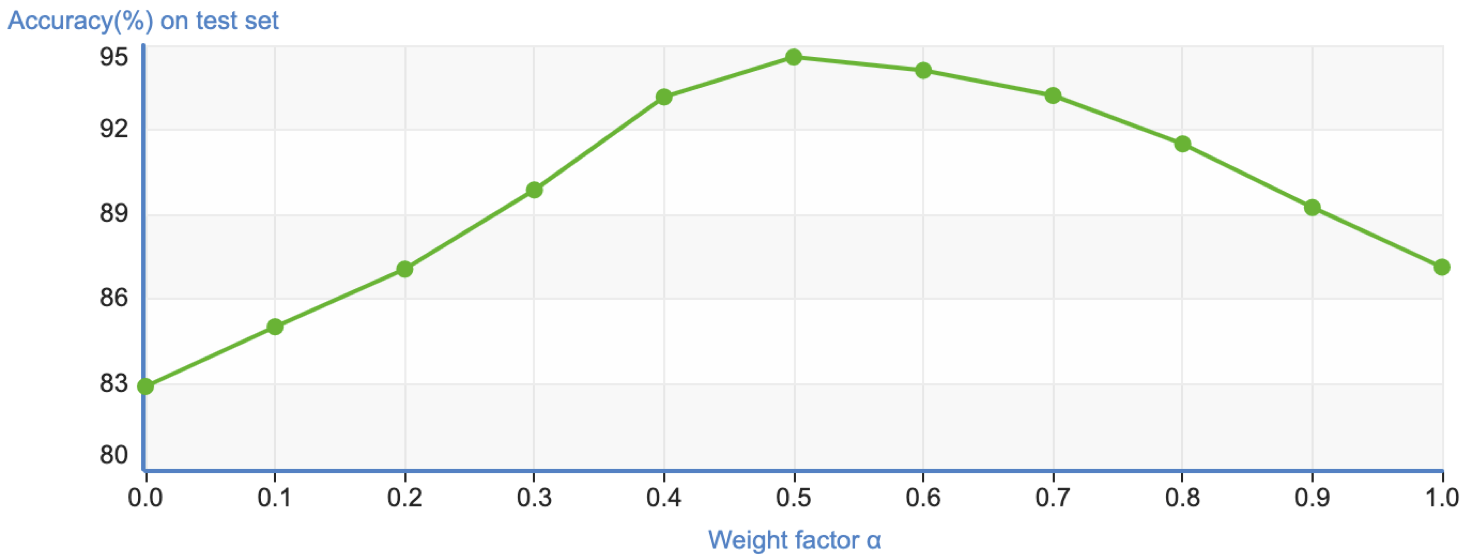}}
  \caption{The influence of the weight factor $\alpha$ to classification accuracy where $\alpha$ ranging from 0 to 1.}
  \label{fig:weight_factor}
\end{minipage}
\end{figure}

\subsubsection{The Influence of the Rectification Module to Convergence Speed}
We take the case under mixed noise with noise level $\rho = 0.2$ and $\rho = 0.3$ as examples to explore the influence of the rectification module to convergence speed. In Fig.~\ref{fig:Progressive}, we show the comparison with respect to classification accuracy and convergence speed between with and without rectification module. It can be found that, when without the rectification module, the training process is unstable; in contrast, when with the rectification module, the training process becomes stable in about $27^{th}$ iteration, and the classification accuracy is also improved.

\subsubsection{The Influence of $Q$ to Classification Performance}
$Q$ is the number of mini-batches $\{( \boldsymbol{X},\hat{\boldsymbol{Y}}^{i}  )\}_{i=1}^Q$ with pseudo labels that we generate for each mini-batch $(\boldsymbol{X},\boldsymbol{Y})$ from the original training data. As shown in Fig.~\ref{fig:parameter q}, we test four cases with $Q=0,5,10,15$ under symmetric noise (SN) and asymmetric noise (AN) with various noise levels. It is worth noting that, when $Q=0$ our scheme becomes Cross-Entropy. It can be seen that the classification accuracy is improved as $Q$ increases. We can infer that, more synthetic mini-batches are involved into self-learning, the model is more likely to learn accurate labels, and the network becomes more robust to noise. 
In practical implementation, we set $Q=10$ to achieve the balance between performance and training speed.

\subsubsection{The Influence of $\alpha$ to Classification Performance}
As shown in Eq. (\ref{eq:alpha}),
$\alpha \in \left [ 0,1 \right ]$ is the weight factor that trade-offs the contributions of the cross-entropy loss and the meta-learning loss. In another word, $\alpha$ decides the network shall focus on the original labels or on the pseudo labels, which greatly affects the classification performance.

We take the case under mixed noise (MN) with noise level $\rho = 0.4$ as an example to study the influence of different $\alpha$ ranging from 0 to 1 with step 0.1 to the classification accuracy.  There exists two extreme cases: (i) When $\alpha=1$, the network is trained by using only pseudo noisy labels without original noisy labels. (ii) When $\alpha=0$, the training process abandons the rectified meta-learning phase and only adopts the primary learning phase with original noisy labels.

From the Fig. \ref{fig:weight_factor}, it can be found that when $\alpha=0$ (without using the corrected pseudo labels) the classification performance is poor; when $\alpha=1$ (using only corrected pseudo labels) the performance is also not satisfactory but higher than the case $\alpha=0$. The accuracy curve shows that the corrected pseudo labels would erroneously correct some clean labels. During the training process, directly using all corrected pseudo labels will bring new noise, and make the network fit on simple features.  The model jointly trained by both the original noisy labels and the corrected pseudo labels achieves the best performance when $\alpha=0.5$.

\section{Conclusion}

In this paper, we presented an effective framework for plant diseases classification that combines rectified meta-learning and a rectification module to learn from noisy labels. Our approach contains two phases, the primary learning phase and the rectified meta-learning phase. The rectified meta-learning phase is tailored to make the primary learning phase less prone to overfitting to label noise. Specifically, based on multiple synthetic mini-batches, we train the network using one gradient update, and enforce it to give consistent predictions with that of primary learning phase. A rectification module is further designed to pay more attention to the unbiased samples, leading to accelerated convergence and improved classification accuracy.  Experimental results validate the superior performance of our method compared to state-of-the-art methods.

\vspace{1cm}
\ifCLASSOPTIONcaptionsoff
  \newpage
\fi



%



\bibliographystyle{IEEEbib}
\bibliography{refs}

%








\end{document}